\documentclass[11pt,a4paper]{article}
\usepackage[hyperref]{acl2018}
\usepackage{times}
\usepackage{latexsym}
\usepackage{amssymb}
\usepackage{url}
\usepackage{graphicx}
\usepackage{amsmath}
\usepackage{breqn}
\usepackage{algorithm}
\usepackage{algorithmicx}  
\usepackage{algpseudocode} 
\usepackage{url}
\usepackage{xcolor}
\usepackage{multirow}
\usepackage{subfigure}

\aclfinalcopy % Uncomment this line for the final submission
\def\aclpaperid{1513} %  Enter the acl Paper ID here

\title{DSGAN: Generative Adversarial Training for Distant Supervision Relation Extraction}

\author{Pengda Qin$^\sharp$, Weiran Xu$^\sharp$, William Yang Wang$^\flat$ \\
  $^\sharp$Beijing University of Posts and Telecommunications, China \\
  $^\flat$University of California, Santa Barbara, USA\\
  {\tt \{qinpengda, xuweiran\}@bupt.edu.cn} \\
  {\tt \{william\}@cs.ucsb.edu} \\
  }

\date{}

\begin{document}
\maketitle
\begin{abstract}
Distant supervision can effectively label data for relation extraction, but suffers from the noise labeling problem. Recent works mainly perform soft bag-level noise reduction strategies to find the relatively better samples in a sentence bag, which is suboptimal compared with making a hard decision of false positive samples in sentence level. In this paper, we introduce an adversarial learning framework, which we named DSGAN, to learn a sentence-level true-positive generator. Inspired by Generative Adversarial Networks, we regard the positive samples generated by the generator as the negative samples to train the discriminator. The optimal generator is obtained until the discrimination ability of the discriminator has the greatest decline. We adopt the generator to filter distant supervision training dataset and redistribute the false positive instances into the negative set, in which way to provide a cleaned dataset for relation classification. The experimental results show that the proposed strategy significantly improves the performance of distant supervision relation extraction comparing to state-of-the-art systems.
%This adversarial training process is independent of relation extraction. 
\end{abstract}

\section{Introduction}
\label{introduction}

\begin{figure}[t]
\begin{center}
\includegraphics[width=8cm]{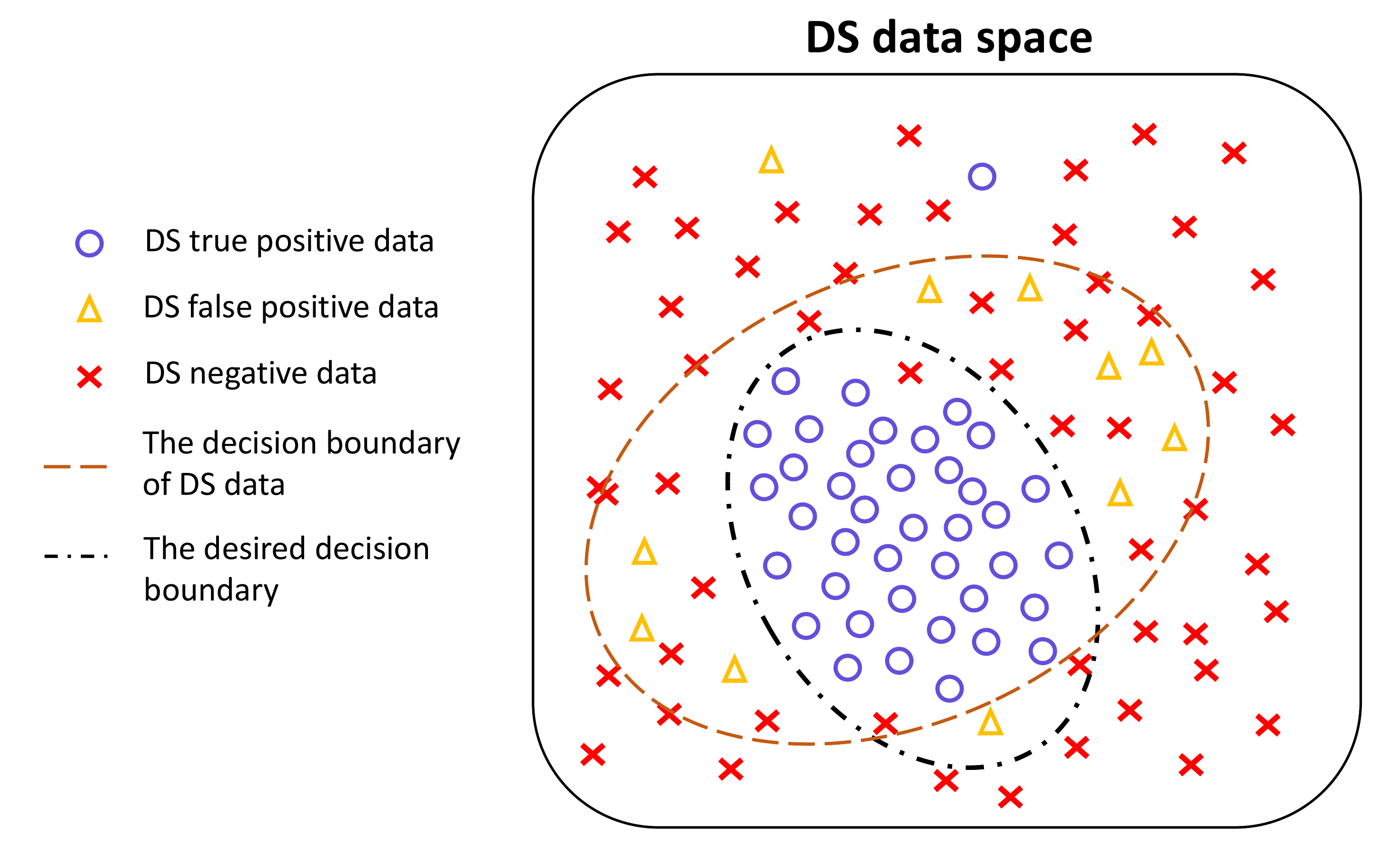}
\caption{Illustration of the distant supervision training data distribution for one relation type.}
\label{fig:data_distribution}
\end{center}
\vspace{-3ex}
\end{figure}

Relation extraction is a crucial task in the field of natural language processing (NLP). It has a wide range of applications including information retrieval, question answering, and knowledge base completion. The goal of relation extraction system is to predict relation between entity pair in a sentence~\cite{zelenko2003kernel,bunescu2005subsequence,guodong2005exploring}. For example, given a sentence ``The ${[owl]}_{e1}$ held the mouse in its ${[claw]}_{e2}$.", a relation classifier should figure out the relation {\bf Component-Whole} between entity ${owl}$ and ${claw}$.

With the infinite amount of facts in real world, it is extremely expensive, and almost impossible for human annotators to annotate training dataset to meet the needs of all walks of life. This problem has received increasingly attention. Few-shot learning and Zero-shot Learning~\cite{xian2017zero} try to predict the unseen classes with few labeled data or even without labeled data. Differently, distant supervision~\cite{mintz2009distant,hoffmann2011knowledge,surdeanu2012multi} is to efficiently generate relational data from plain text for unseen relations with distant supervision (DS). However, it naturally brings with some defects: the resulted distantly-supervised training samples are often very noisy (shown in Figure~\ref{fig:data_distribution}), which is the main problem of impeding the performance~\cite{roth2013survey}.
Most of the current state-of-the-art methods~\cite{zeng2015distant, lin2016neural} make the denoising operation in the sentence bag of entity pair, and integrate this process into the distant supervision relation extraction. Indeed, these methods can filter a substantial number of noise samples; However, they overlook the case that all sentences of an entity pair are false positive, which is also the common phenomenon in distant supervision datasets. Under this consideration, an independent and  accurate \textbf{sentence-level} noise reduction strategy is the better choice. 

In this paper, we design an adversarial learning process~\cite{goodfellow2014generative,radford2015unsupervised} to obtain a sentence-level generator that can recognize the true positive samples from the noisy distant supervision dataset without any supervised information. In Figure~\ref{fig:data_distribution}, the existence of false positive samples makes the DS decision boundary suboptimal, therefore hinders the performance of relation extraction. However, in terms of quantity, the true positive samples still occupy most of the proportion; this is the prerequisite of our method. Given the discriminator that possesses the decision boundary of DS dataset (the brown decision boundary in Figure~\ref{fig:data_distribution}), the generator tries to generate true positive samples from DS positive dataset;
Then, we assign the generated samples with negative label and the rest samples with positive label to challenge the discriminator. Under this adversarial setting, if the generated sample set includes more true positive samples and more false positive samples are left in the rest set, the classification ability of the discriminator will drop faster. Empirically, we show that our method has brought consistent performance gains in various deep-neural-network-based models, achieving strong performances on the widely used New York Times dataset~\cite{riedel2010modeling}.
Our contributions are three-fold:
\begin{itemize}
\item We are the first to consider adversarial learning to denoise the distant supervision relation extraction dataset.
\item Our method is sentence-level and model-agnostic, so it can be used as a plug-and-play technique for any relation extractors.
\item We show that our method can generate a cleaned dataset without any supervised information, in which way to boost the performance of recently proposed neural relation extractors.
\end{itemize}

In Section~\ref{sec:related}, we outline some related works on distant supervision relation extraction. Next, we describe our adversarial learning strategy in Section~\ref{sec:method}. In Section~\ref{sec:exp}, we show the stability analyses of DSGAN and the empirical evaluation results. And finally, we conclude in Section~\ref{sec:conclude}.

\section{Related Work}
\label{sec:related}

To address the above-mentioned data sparsity issue, \newcite{mintz2009distant} first align unlabeled text corpus with Freebase by distant supervision. However, distant supervision inevitably suffers from the wrong labeling problem. Instead of explicitly removing noisy instances, the early works intend to suppress the noise. \newcite{riedel2010modeling} adopt multi-instance single-label learning in relation extraction; \newcite{hoffmann2011knowledge} and \newcite{surdeanu2012multi} model distant supervision relation extraction as a multi-instance multi-label problem. 

Recently, some deep-learning-based models~\cite{zeng2014relation,shen-huang:2016:COLING} have been proposed to solve relation extraction.
Naturally, some works try to alleviate the wrong labeling problem with deep learning technique, and their denoising process is integrated into relation extraction. \newcite{zeng2015distant} select one most plausible sentence to represent the relation between entity pairs, which inevitably misses some valuable information. ~\newcite{lin2016neural} calculate a series of soft attention weights for all sentences of one entity pair and the incorrect sentences can be down-weighted; Base on the same idea, \newcite{ji2017distant} bring the useful entity information into the calculation of the attention weights. However, compared to these soft attention weight assignment strategies, recognizing the true positive samples from distant supervision dataset before relation extraction is a better choice. \newcite{takamatsu2012reducing} build a noise-filtering strategy based on the linguistic features extracted from many NLP tools, including NER and dependency tree, which inevitably suffers the error propagation problem; while we just utilize word embedding as the input information. In this work, we learn a true-positive identifier (the generator) which is independent of the relation prediction of entity pairs, so it can be directly applied on top of any existing relation extraction classifiers. Then, we redistribute the false positive samples into the negative set, in which way to make full use of the distantly labeled resources.

\section{Adversarial Learning for Distant Supervision}
\label{sec:method}

\begin{figure*}[t]
\begin{center}
\includegraphics[width=16cm]{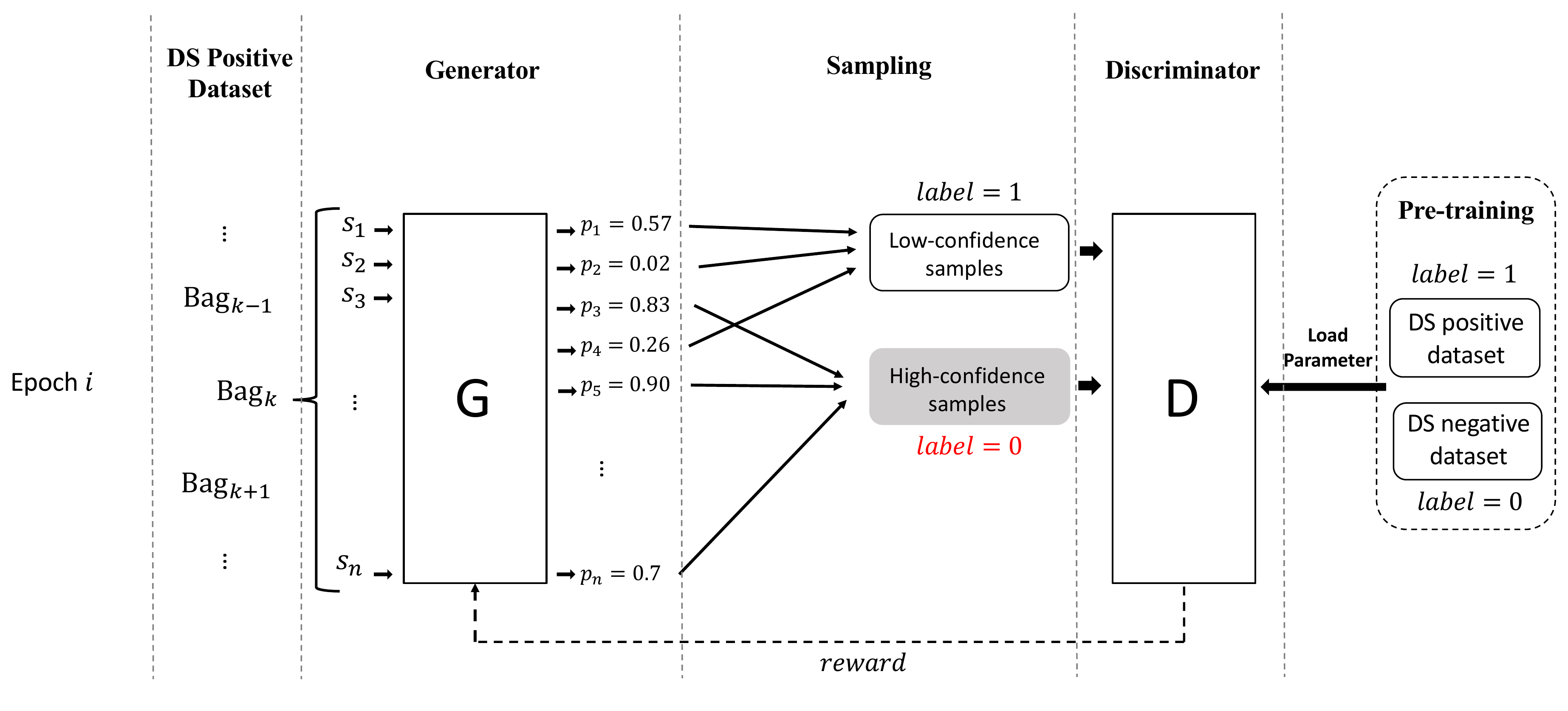}
\caption{\label{framework} An overview of the DSGAN training pipeline. 
%In each epoch, the discriminator (denoted D) loads the same set of pre-trained parameter. 
The generator (denoted by {\bf G}) calculates the probability distribution over a bag of DS positive samples, and then samples according to this probability distribution. The high-confidence samples generated by G are regarded as true positive samples.
The discriminator (denoted by {\bf D}) receives these high-confidence samples but regards them as negative samples; conversely, the low-confidence samples are still treated as positive samples. For the generated samples, {\bf G} \textbf{maximizes} the probability of being true positive; on the contrary, {\bf D} \textbf{minimizes} this probability.}
\end{center}
\end{figure*}

In this section, we introduce an adversarial learning pipeline to obtain a robust generator which can automatically discover the true positive samples from the noisy distantly-supervised dataset without any supervised information. 
The overview of our adversarial learning process is shown in Figure~\ref{framework}.
Given a set of distantly-labeled sentences, the generator tries to generate true positive samples from it; But, these generated samples are regarded as negative samples to train the discriminator. Thus, when finishing scanning the DS positive dataset one time, the more true positive samples that the generator discovers, the sharper drop of performance the discriminator obtains. After adversarial training, we hope to obtain a robust generator that is capable of forcing discriminator into maximumly losing its classification ability.

In the following section, we describe the adversarial training pipeline between the generator and the discriminator, including the pre-training strategy, objective functions and gradient calculation. Because the generator involves a discrete sampling step, we introduce a policy gradient method to calculate gradients for the generator. 

\subsection{Pre-Training Strategy}
Both the generator and the discriminator require the pre-training process, which is the common setting for GANs~\cite{cai2017kbgan,wang2017irgan}. With the better initial parameters, the adversarial learning is prone to convergence. As presented in Figure~\ref{framework}, the discriminator is pre-trained with DS positive dataset $P$ (label 1) and DS negative set $N^D$ (label 0). After our adversarial learning process, we desire a strong generator that can, to the maximum extent, collapse the discriminator. Therefore, the more robust generator can be obtained via competing with the more robust discriminator. So we pre-train the discriminator until the accuracy reaches 90\% or more. The pre-training of generator is similar to the discriminator; however, for the negative dataset, we use another completely different dataset $N^G$, which makes sure the robustness of the experiment. Specially, we let the generator overfits the DS positive dataset $P$. The reason of this setting is that we hope the generator wrongly give high probabilities to all of the noisy DS positive samples at the beginning of the training process. Then, along with our adversarial learning, the generator learns to gradually decrease the probabilities of the false positive samples. 

\subsection{Generative Adversarial Training for Distant Supervision Relation Extraction}

%In parallel to terminologies used in GAN literature, we will simply call these two models \emph{generator} and \emph{discriminator} respectively in the rest of this paper. 
The generator and the discriminator of DSGAN are both modeled by simple CNN, because CNN performs well in understanding sentence~\cite{zeng2014relation}, and it has less parameters than RNN-based networks. For relation extraction, the input information consists of the sentences and entity pairs; thus, as the common setting~\cite{zeng2014relation,nguyen2015event}, we use both word embedding and position embedding to convert input instances into continuous real-valued vectors.

What we desire the generator to do is to accurately recognize true positive samples. Unlike the generator applied in computer vision field~\cite{im2016generating} that generates new image from the input noise, our generator just needs to discover true positive samples from the noisy DS positive dataset. Thus, it is to realize the ``sampling from a probability distribution'' process of the discrete GANs (Figure~\ref{framework}). For a input sentence $s_j$, we define the probability of being true positive sample by generator as $p_G(s_j)$. Similarly, for discriminator, the probability of being true positive sample is represented as $p_D(s_j)$. We define that one epoch means that one time scanning of the entire DS positive dataset. In order to obtain more feedbacks and make the training process more efficient, we split the DS positive dataset $P=\{s_1,s_2,...,s_j,...\}$ into $N$ bags $B=\{B^1, B^2,...B^N\}$, and the network parameters $\theta_G$, $\theta_D$ are updated when finishing processing one bag $B_i$\footnote{The \emph{bag} here has the different definition from the sentence \emph{bag} of an entity pair mentioned in the Section~\ref{introduction}.}. Based on the notion of adversarial learning, we define the objectives of the generator and the discriminator as follow, and they are alternatively trained towards their respective objectives.
%What needs to be explained is that the input sequence of these bags is constant, which means the bag set $B$ is identical for each epoch.

\renewcommand{\algorithmicrequire}{\textbf{Input:}} % Use Input in the format of Algorithm
\renewcommand{\algorithmicensure}{\textbf{Output:}} % Use Output in the format of Algorithm
\renewcommand{\algorithmiclastcon}{\textbf{Data:}}

 \begin{algorithm*}[t]  
        \caption{The DSGAN algorithm.}  
        \begin{algorithmic}[1]  
            \lastcon DS positive set $\mathnormal{P}$, DS negative set $\mathnormal{N^G}$ for generator G, DS negative set $\mathnormal{N^D}$ for discriminator D%threshold $\gamma_k$
            \Require Pre-trained G with parameters $\mathnormal{\theta_G}$ on dataset ($\mathnormal{P}$, $\mathnormal{N^G}$); Pre-trained D with parameters $\mathnormal{\theta_D}$ on dataset ($\mathnormal{P}$, $\mathnormal{N^D}$)
            \Ensure Adversarially trained generator G
            \State Load parameters $\mathnormal{\theta_G}$ for G
            \State Split $\mathnormal{P}$ into the bag sequence $P=\{{B^1},{B^2},...,{B^i},...,{B^N}\}$
            %\For{epoch $l = 1 \to L$}
            \Repeat
            	\State Load parameters $\mathnormal{\theta_D}$ for D
            	\State ${G_G} \leftarrow 0, {G_D} \leftarrow 0$
                %\State Compute the average negative label probability of $N^D$ with current $\theta_D$. 
                %\State $p_{N^D} = {\frac{1}{|N^D|}}\sum_{s_j \in N^D} {p_D}(0|s_j)$
                %\State Compute the accuracy ${ACC}_0$ of $N_D$ with current $\theta_D$
                \For{$B_i \in \mathnormal P, i=1 \, $to$ \, N$}
                	\State Compute the probability $p_G(s_j)$ for each sentence $s_j$ in $B_i$
                	\State Obtain the generated part $T$ by sampling according to ${\{p_G(s_j)\}}_{j=1...|B|}$ and the rest set $F=B_i-T$
                    \State $G_D \leftarrow - \frac{1}{|P|} \{ {\bigtriangledown_{\theta_D}}\sum^{T}{\log (1-p_D(s_j))} + {\bigtriangledown_{\theta_D}}\sum^{F}{\log p_D(s_j)} \}$
                    \State $\theta_D \leftarrow {\theta}_D - {\alpha_D} {G_D}$
                    %\State Compute $p_i^{avg}$ and ${ACC}_i$ with current ${\theta}_D$
                    \State Calculate the reward $r$
                    \State $G_G \leftarrow \frac{1}{|T|} \sum^{T} r {\bigtriangledown_{\theta_G}} \log p_G(s_j)$
                    \State ${\theta}_G \leftarrow {\theta}_G + {\alpha_G} G_G$
                \EndFor
                \State Compute the accuracy $ACC_D$ on $N^D$ with the current $\theta_D$
            \Until{$ACC_D$ no longer drops}
            \State Save ${\theta}_G$
        \end{algorithmic}  
    \end{algorithm*}

\paragraph{Generator}
Suppose that the generator produces a set of probability distribution $\{p_G(s_j)\}_{j=1...|B_i|}$ for a sentence bag $B_i$. Based on these probabilities, a set of sentence are sampled and we denote this set as $T$. 

\begin{equation}
T = \{s_j\}, {s_j} \thicksim {p_G}(s_j), j=1,2,...,|B_i|
\label {fake_data}
\end{equation}
This generated dataset $T$ consists of the high-confidence sentences, and is regard as true positive samples by the current generator; however, it will be treated as the negative samples to train the discriminator. In order to challenge the discriminator, the objective of the generator can be formulated as \textbf{maximizing} the following probabilities of the generated dataset $T$:

\begin{equation}
\mathnormal{L_G} = \sum_{{s_j}\in {T}} \log {p_D(s_j)}
\label {G_objective}
\end{equation}
%,  {s_j} \thicksim {p_G}(s_j)} \log {p_D(s_j)
Because $L_G$ involves a discrete sampling step, so it cannot be directly optimized by gradient-based algorithm. We adopt a common approach: the policy-gradient-based reinforcement learning. The following section will give the detailed introduction of the setting of reinforcement learning. The parameters of the generator are continually updated until reaching the convergence condition.

\paragraph{Discriminator}
After the generator has generated the sample subset $T$ , the discriminator treats them as the negative samples; conversely, the rest part $F=B_i-T$ is treated as positive samples. So, the objective of the discriminator can be formulated as \textbf{minimizing} the following cross-entropy loss function:

\begin{dmath}
\mathnormal{L_D} =  -(\sum_{s_j \in (B_i-T)} \log {p_D(s_j)}
+ \sum_{s_j \in T} \log (1-{p_D(s_j)}))
\label {D_objective}
\end{dmath}
The update of discriminator is identical to the common binary classification problem. Naturally, it can be simply optimized by any gradient-based algorithm.

What needs to be explained is that, unlike the common setting of discriminator in previous works, our discriminator \emph{loads the same pre-trained parameter set at the beginning of each epoch} as shown in Figure~\ref{framework}. There are two reasons. First, at the end of our adversarial training, what we need is a robust generator rather than a discriminator. Second, our generator is to sample data rather than generate new data from scratch; Therefore, the discriminator is relatively easy to be collapsed. So we design this new adversarial strategy: the robustest generator is yielded when the discriminator has the largest drop of performance in one epoch. In order to create the equal condition, the bag set $B$ for each epoch is identical, including the sequence and the sentences in each bag $B_i$.

\paragraph{Optimizing Generator}
The objective of the generator is similar to the objective of the one-step reinforcement learning problem: Maximizing the expectation of a given function of samples from a parametrized probability distribution. Therefore, we use a policy gradient strategy to update the generator. Corresponding to the terminology of reinforcement learning, $s_j$ is the \emph{state} and $P_G(s_j)$ is the \emph{policy}. In order to better reflect the quality of the generator, we define the reward $r$ from two angles:

\begin{itemize}
\item As the common setting in adversarial learning, for the generated sample set, we hope the confidence of being positive samples by the discriminator becomes higher. Therefore, the first component of our reward is formulated as below:

\begin{equation}
{r_1} = {\frac{1}{|T|}}\sum_{s_j \in T} {p_D}(s_j) - b_1
\label {reward1}
\end{equation}
the function of $b_1$ is to reduce variance during reinforcement learning. 
%$b_1$ is the average probability of $T$ calculated by the pre-trained discriminator.

\item The second component is from the average prediction probability of $N^D$,
\begin{equation}
\tilde{p} = {\frac{1}{|N^D|}}\sum_{s_j \in N^D} {p_D}(s_j)
\label {P_avg}
\end{equation}
$N^D$ participates the pre-training process of the discriminator, but not the adversarial training process. When the classification capacity of discriminator declines, the accuracy of being predicted as negative sample on $N^D$ gradually drops; thus, $\tilde{p}$ increases. In other words, the generator becomes better. Therefore, for epoch $k$, after processing the bag $B_i$, reward $r_2$ is calculated as below,

\begin{equation}
\begin{split}
& \qquad \qquad \quad {r_2} = \eta({{\tilde{p}}_i^k} - b_2) \\
& where \,\, b_2\!=\!\max \{{{\tilde{p}}_i^m}\}, m\!=\!1...,k\!-\!1 
\end{split}
\label {reward2}
\end{equation}
%where $\eta$ is the scaling coefficient. 
$b_2$ has the same function as $b_1$.

\end{itemize}

The gradient of $L_G$ can be formulated as below:

\begin{equation}
\begin{split}
{\bigtriangledown_{\theta_D}}{L_G} &= \sum_{s_j \in B_i} \mathbb{E}_{s_j \thicksim p_G(s_j)}  r {\bigtriangledown_{\theta_G}} \log p_G(s_j) \\
&= \frac{1}{|T|} \sum_{s_j \in T} r {\bigtriangledown_{\theta_G}} \log p_G(s_j)
\end{split}
\label {G_update}
\end{equation}

\subsection{Cleaning Noisy Dataset with Generator}
\label{ssec:redistribute}

After our adversarial learning process, we obtain one generator for one relation type; These generators possess the capability of generating true positive samples for the corresponding relation type. Thus, we can adopt the generator to filter the noise samples from distant supervision dataset. Simply and clearly, we utilize the generator as a binary classifier. In order to reach the maximum utilization of data, we develop a strategy: for an entity pair with a set of annotated sentences, if all of these sentences are determined as false negative by our generator, this entity pair will be redistributed into the negative set. Under this strategy, the scale of distant supervision training set keeps unchanged.

\section{Experiments}
\label{sec:exp}
This paper proposes an adversarial learning strategy to detect true positive samples from the noisy distant supervision dataset. Due to the absence of supervised information, we define a generator to heuristically learn to recognize true positive samples through competing with a discriminator. Therefore, our experiments are intended to demonstrate that our DSGAN method possess this capability. To this end, we first briefly introduce the dataset and the evaluation metrics. Empirically, the adversarial learning process, to some extent, has instability; Therefore, we next illustrate the convergence of our adversarial training process. Finally, we demonstrate the efficiency of our generator from two angles: the quality of the generated samples and the performance on the widely-used distant supervision relation extraction task.

\begin{figure*}[t]
\begin{center}
\includegraphics[width=16cm]{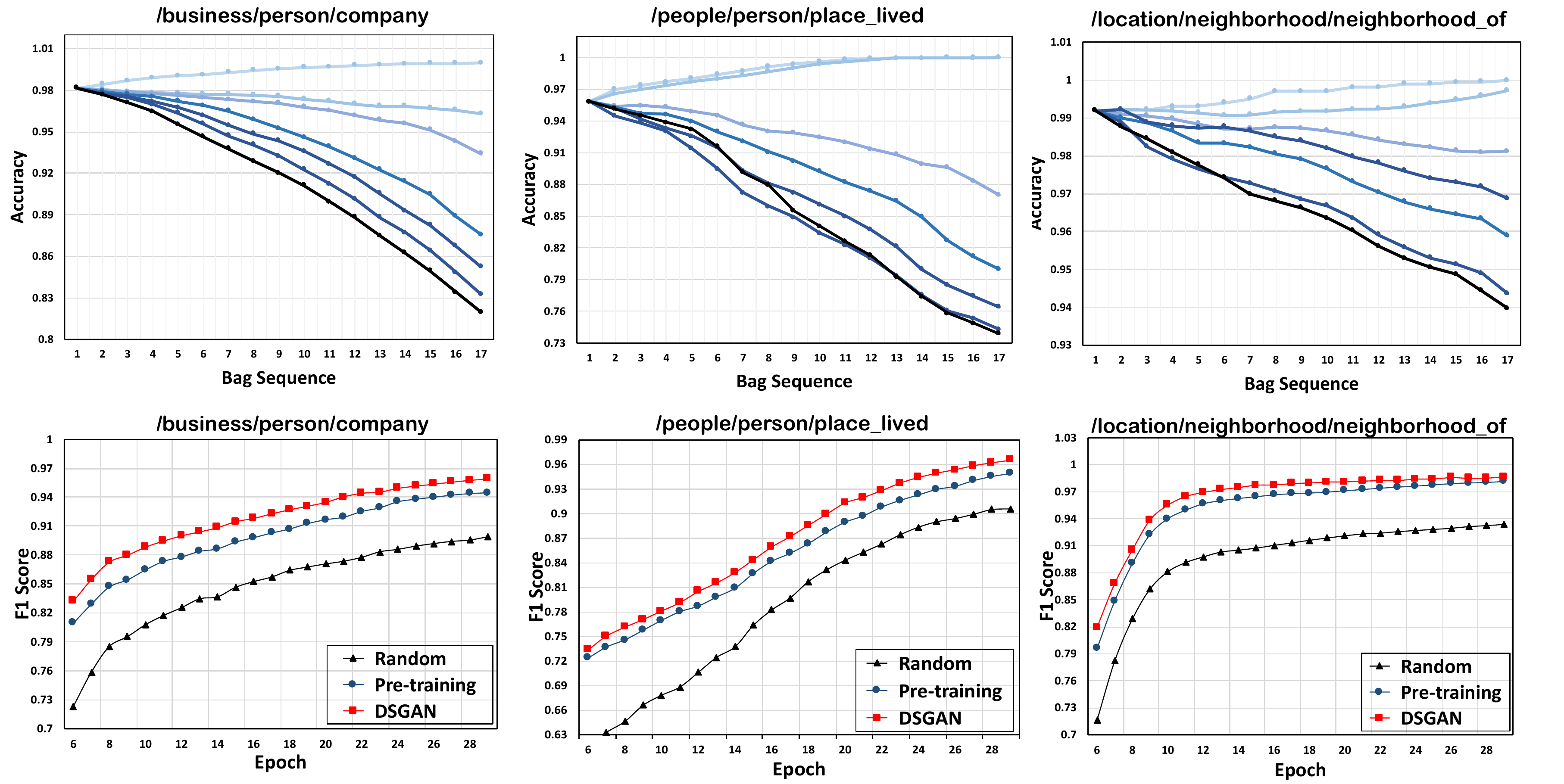}
\caption{The convergence of the DSGAN training process for 3 relation types and the performance of their corresponding generators. The figures in the first row present the performance change on $N^D$ in some specific epochs during processing the $B=\{B^1, B^2,...B^N\}$. Each curve stands for one epoch; The color of curves become darker as long as the epoch goes on. Because the discriminator reloads the pre-trained parameters at the beginning of each epoch, all curves start from the same point for each relation type; Along with the adversarial training, the generator gradually collapses the discriminator. The figures in the second row reflect the performance of generators from the view of the difficulty level of training with the positive datasets that are generated by different strategies. Based on the noisy DS positive dataset $P$, \emph{DSGAN} represents that the cleaned positive dataset is generated by our DSGAN generator; \emph{Random} means that the positive set is randomly selected from $P$; \emph{Pre-training} denotes that the dataset is selected according to the prediction probability of the pre-trained generator. These three new positive datasets are in the same size.}
\label{fig:all_figures}
\end{center}
\vspace{-3ex}
\end{figure*}

\subsection{Evaluation and Implementation Details}
The Reidel dataset\footnote{http://iesl.cs.umass.edu/riedel/ecml/}~\cite{riedel2010modeling} is a commonly-used distant supervision relation extraction dataset. Freebase is a huge knowledge base including billions of triples: the entity pair and the specific relationship between them. Given these triples, the sentences of each entity pair are selected from the New York Times corpus(NYT). Entity mentions of NYT corpus are recognized by the Stanford named entity recognizer~\cite{finkel2005incorporating}. 
There are 52 actual relationships and a special relation $NA$ which indicates there is no relation between head and tail entities. Entity pairs of $NA$ are defined as the entity pairs that appear in the same sentence but are not related according to Freebase.

Due to the absence of the corresponding labeled dataset, there is not a ground-truth test dataset to evaluate the performance of distant supervision relation extraction system. 
Under this circumstance, the previous work adopt the held-out evaluation to evaluate their systems, which can provide an approximate measure of precision without requiring costly human evaluation. It builds a test set where entity pairs are also extracted from Freebase. Similarly, relation facts that discovered from test articles are automatically compared with those in Freebase. CNN is widely used in relation classification~\cite{santos2015classifying,qin2017designing}, thus the generator and the discriminator are both modeled as a simple CNN with the window size $c_w$ and the kernel size $c_k$. Word embedding is directly from the released word embedding matrix by ~\newcite{lin2016neural}\footnote{https://github.com/thunlp/NRE}. Position embedding has the same setting with the previous works: the maximum distance of -30 and 30. Some detailed hyperparameter settings are displayed in Table~\ref{paraset}.

\begin{table}[t]
\normalsize
\begin{tabular}{cc}
\hline
\hline 
\bf {Hyperparameter} & \bf {Value} \\
\hline
CNN Window $c_w$,\, kernel size $c_k$ & 3,\,\,100  \\
Word embedding $d_e$, $|V|$ & 50, 114042  \\
Position embedding $d_p$& 5\\
Learning rate of G,\,D & 1e-5,\,\,1e-4\\
%${\eta}_1$,\,${\eta}_2$ & 20,\,\,30\\
\hline
\hline
\end{tabular}
\caption{\label{paraset} Hyperparameter settings of the generator and the discriminator.}
\end{table}

\subsection{Training Process of DSGAN}

\begin{figure}[!htp]
\centering
\begin{center}
\includegraphics[width=7cm]{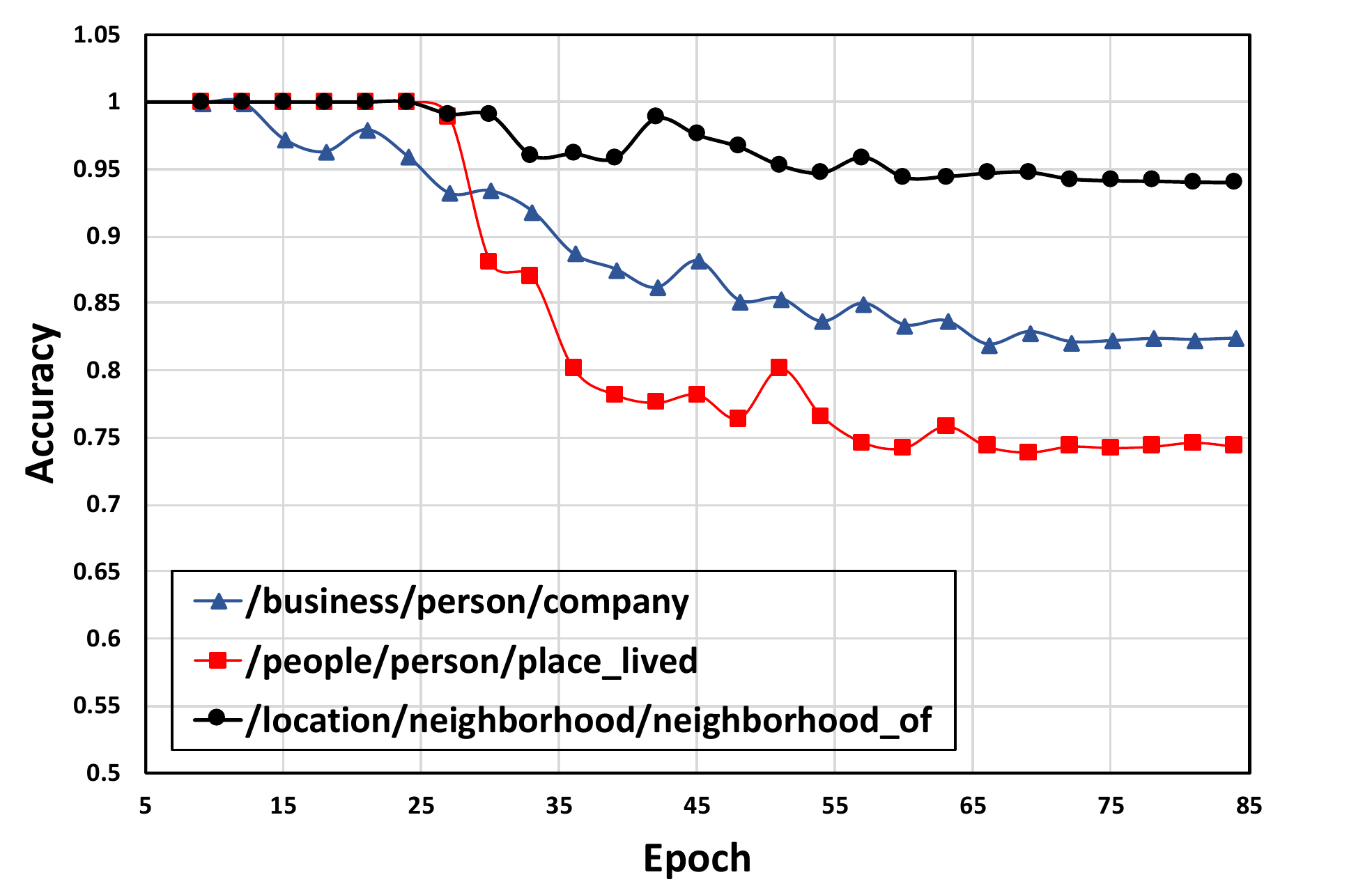}
\caption{The performance change of the discriminator on $N^D$ during the training process. Each point in the curves records the prediction accuracy on $N^D$ when finishing each epoch. We stop the training process when this accuracy no longer decreases.}
\label{fig:neg_acc}
\end{center}
\vspace{-3ex}
\end{figure}

Because adversarial learning is widely regarded as an effective but unstable technique, here we illustrate some property changes during the training process, in which way to indicate the learning trend of our proposed approach. We use 3 relation types as the examples: \emph{/business/person/company}, \emph{/people/person/place\_lived} and \emph{/location/neighborhood/neighborhood\_of}. Because they are from three major classes (\emph{bussiness}, \emph{people}, \emph{location}) of Reidel dataset and they all have enough distant-supervised instances. The first row in Figure~\ref{fig:all_figures} shows the classification ability change of the discriminator during training. The accuracy is calculated from the negative set\footnote{The trends in the first row of Figure~\ref{fig:all_figures} is not limited in $N_D$. Different randomly-selected negative sets have the same trends.} $N^D$. At the beginning of adversarial learning, the discriminator performs well on $N^D$; moreover, $N^D$ is not used during adversarial training. Therefore, the accuracy on $N^D$ is the criterion to reflect the performance of the discriminator.
%As can be seen, with the same pre-trained parameters, the performances of discriminator are identical at the beginning of each epoch. 
In the early epochs, the generated samples from the generator increases the accuracy, because it has not possessed the ability of challenging the discriminator; however, as the training epoch increases, this accuracy gradually decreases, which means the discriminator becomes weaker. It is because the generator gradually learn to generate more accurate true positive samples in each bag. After the proposed adversarial learning process, the generator is strong enough to collapse the discriminator. Figure~\ref{fig:neg_acc} gives more intuitive display of the trend of accuracy. Note that there is a critical point of the decline of accuracy for each presented relation types. It is because that the chance we give the generator to challenge the discriminator is just one time scanning of the noisy dataset; this critical point is yielded when the generator has already been robust enough. Thus, we stop the training process when the model reaches this critical point. To sum up, the capability of our generator can steadily increases, which indicates that DSGAN is a robust adversarial learning strategy.

\subsection{Quality of Generator}

Due to the absence of supervised information, we validate the quality of the generator from another angle. Combining with Figure~\ref{fig:data_distribution}, for one relation type, the true positive samples must have evidently higher relevance (the cluster of purple circles). Therefore, a positive set with more true positive samples is easier to be trained; In other words, the convergence speed is faster and the fitting degree on training set is higher. Based on this , we present the comparison tests in the second row of Figure~\ref{fig:all_figures}. We build three positive datasets from the noisy distant supervision dataset $P$: the randomly-selected positive set, the positive set base on the pre-trained generator and the positive set base on the DSGAN generator. For the pre-trained generator, the positive set is selected according to the probability of being positive from high to low. These three sets have the same size and are accompanied by the same negative set. Obviously, the positive set from the DSGAN generator yields the best performance, which indicates that our adversarial learning process is able to produce a robust true-positive generator. In addition, the pre-trained generator also has a good performance; however, compared with the DSGAN generator, it cannot provide the boundary between the false positives and the true positives.

\subsection{Performance on Distant Supervision Relation Extraction}
\begin{figure}[t]
\begin{center}
\includegraphics[width=7cm]{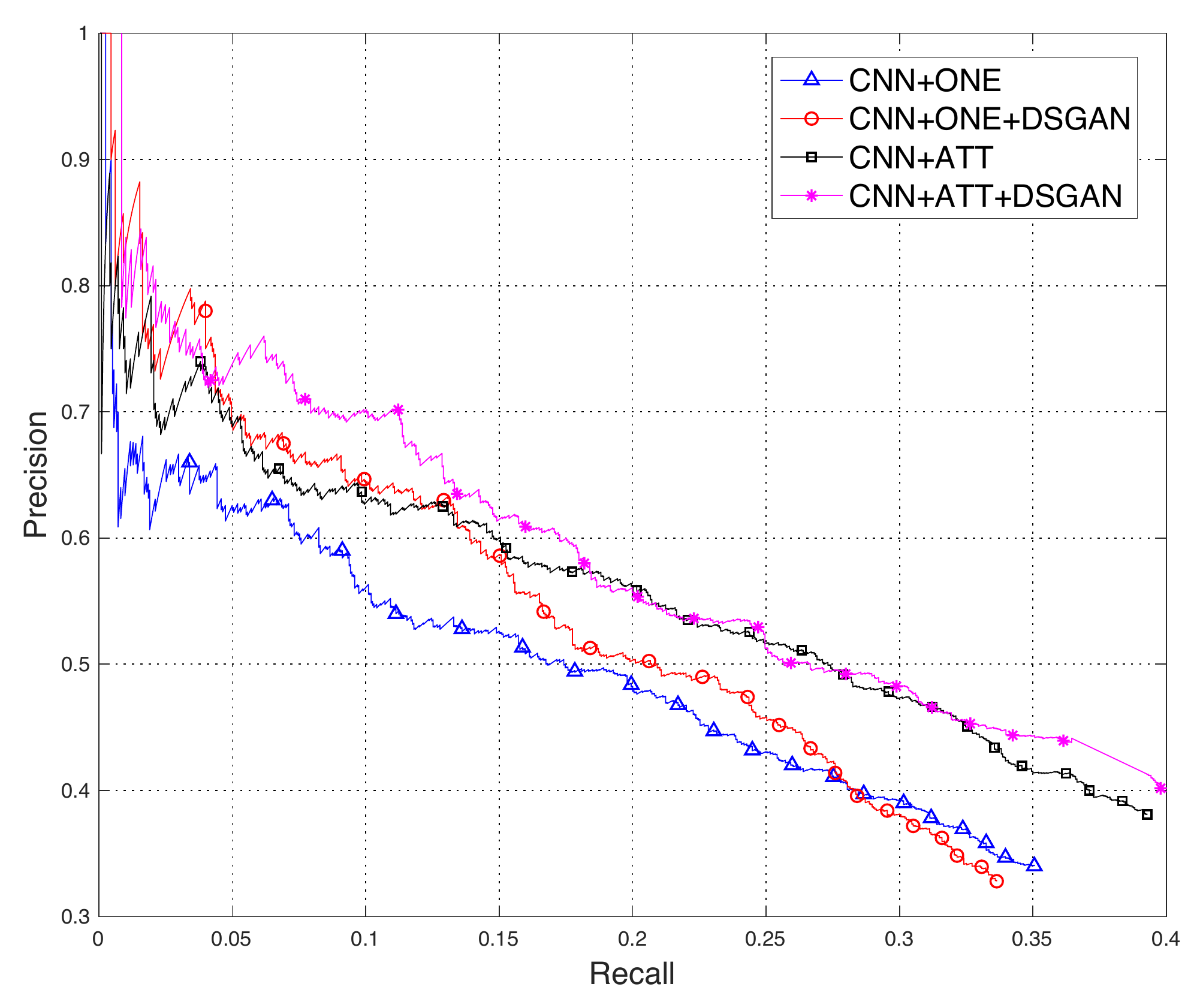}
\caption{\label{CNN_curve}Aggregate PR curves of CNN_based model.}
\end{center}
\end{figure}

\begin{figure}[t]
\begin{center}
\includegraphics[width=7cm]{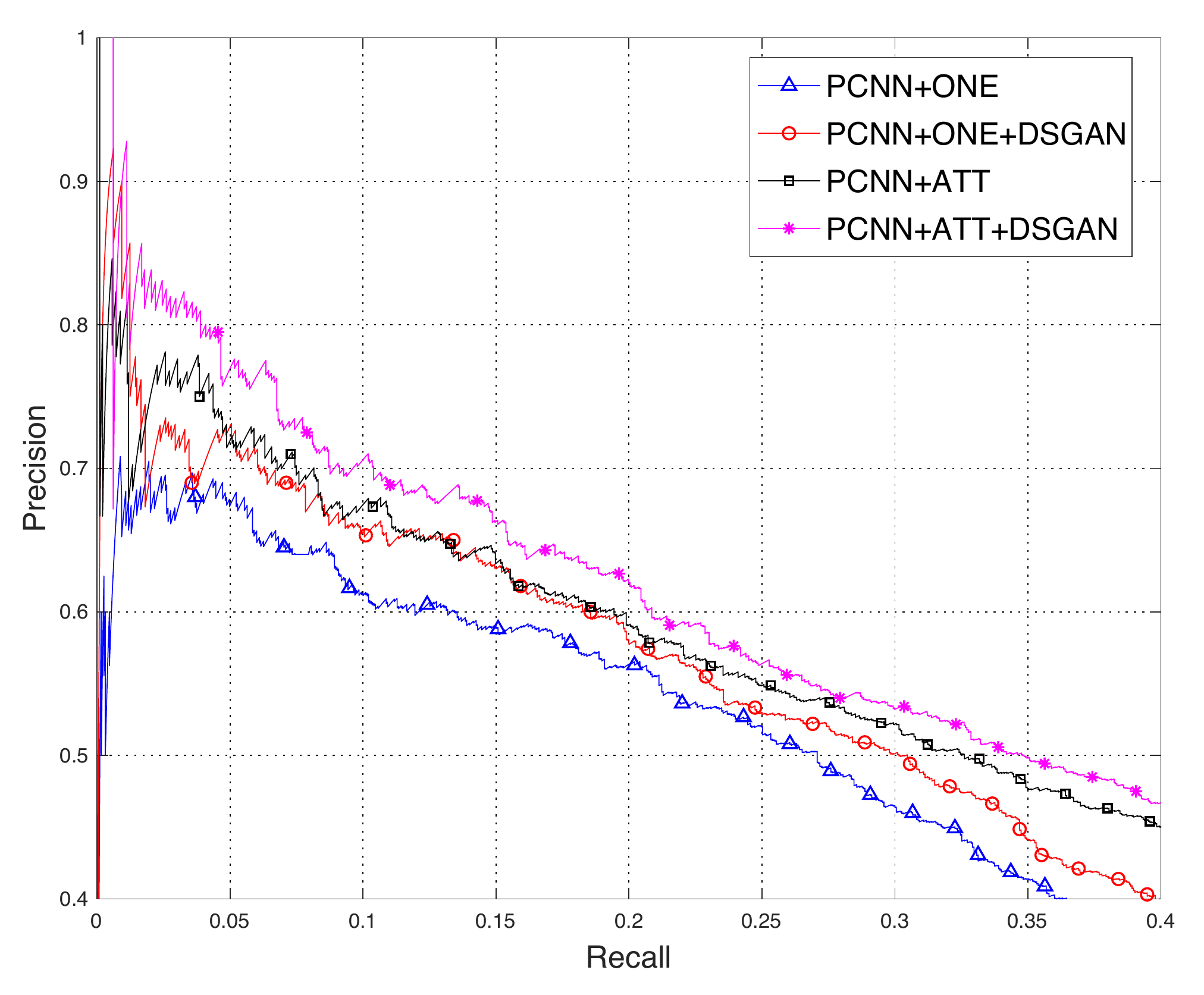}
\caption{\label{PCNN_curve}Aggregate PR curves of PCNN_based model.}
\end{center}
\end{figure}

Based on the proposed adversarial learning process, we obtain a generator that can recognize the true positive samples from the noisy distant supervision dataset. Naturally, the improvement of distant supervision relation extraction can provide a intuitive evaluation of our generator. We adopt the strategy mentioned in Section~\ref{ssec:redistribute} to relocate the dataset. After obtaining this redistributed dataset, we apply it to train the recent state-of-the-art models and observe whether it brings further improvement for these systems. ~\newcite{zeng2015distant} and~\newcite{lin2016neural} are both the robust models to solve wrong labeling problem of distant supervision relation extraction. According to the comparison displayed in Figure~\ref{CNN_curve} and Figure~\ref{PCNN_curve}, all four models (\emph{CNN+ONE}, \emph{CNN+ATT}, \emph{PCNN+ONE} and \emph{PCNN+ATT}) achieve further improvement.

\begin{table}[t]
\normalsize
\centering
\begin{tabular}{lccc}
\hline
\hline 
\bf {Model} & \bf {-} & \bf {+DSGAN} & \bf {p-value} \\
\hline
CNN+ONE & 0.177 & \bf {0.189} & 4.37e-04\\
CNN+ATT & 0.219 & \bf {0.226}  & 8.36e-03 \\
PCNN+ONE & 0.206 &\bf {0.221} & 2.89e-06 \\
PCNN+ATT & 0.253 & \bf {0.264} & 2.34e-03 \\
\hline
\hline
\end{tabular}
\caption{\label{AUC} Comparison of AUC values between previous studies and our DSGAN method. The \emph{p-value} stands for the result of t-test evaluation.}
\end{table}

Even though ~\newcite{zeng2015distant} and ~\newcite{lin2016neural} are designed to alleviate the influence of false positive samples, both of them merely focus on the noise filtering in the sentence bag of entity pairs. ~\newcite{zeng2015distant} combine at-least-one multi-instance learning with deep neural network to extract only one active sentence to represent the target entity pair;~\newcite{lin2016neural} assign soft attention weights to the representations of all sentences of one entity pair, then employ the weighted sum of these representations to predict the relation between the target entity pair. However, from our manual inspection of Riedel dataset~\cite{riedel2010modeling}, we found another false positive case that all the sentences of a specific entity pair are wrong; but the aforementioned methods overlook this case, while the proposed method can solve this problem. Our DSGAN pipeline is independent of the relation prediction of entity pairs, so we can adopt our generator as the true-positive indicator to filter the noisy distant supervision dataset before relation extraction, which explains the origin of these further improvements in Figure~\ref{CNN_curve} and Figure~\ref{PCNN_curve}. In order to give more intuitive comparison, in Table~\ref{AUC}, we present the AUC value of each PR curve, which reflects the area size under these curves. The larger value of AUC reflects the better performance. Also, as can be seen from the result of t-test evaluation, all the p-values are less than 5e-02, so the improvements are obvious.

\section{Conclusion}
\label{sec:conclude}
Distant supervision has become a standard method in relation extraction. However, while it brings the convenience, it also introduces noise in distantly labeled sentences. In this work, we propose the first generative adversarial training method for robust distant supervision relation extraction. More specifically, our framework has two components: a generator that generates true positives, and a discriminator that tries to classify positive and negative data samples. With adversarial training, our goal is to gradually decrease the performance of the discriminator, while the generator improves the performance for predicting true positives when reaching equilibrium. Our approach is model-agnostic, and thus can be applied to any distant supervision model. Empirically, we show that our method can significantly improve the performances of many competitive baselines on the widely used New York Time dataset.

\section*{Acknowledge}
This work was supported by National Natural Science Foundation of China (61702047), Beijing Natural Science Foundation (4174098), the Fundamental Research Funds for the Central Universities (2017RC02) and National Natural Science Foundation of China (61703234)

%\bibliography{acl2018}
\bibliography{acl2018}
\bibliographystyle{acl_natbib}

\end{document}